\documentclass[review]{elsarticle}

\usepackage{lineno,hyperref}
\usepackage{amsmath}
\usepackage{amssymb}
\usepackage{algorithm}
\usepackage{algpseudocode}
\usepackage{subcaption}
\usepackage{tikz}
\usepackage{gensymb}

\modulolinenumbers[5]
\DeclareMathOperator*{\argmax}{arg\,max}

\journal{Acta Astronautica}

\begin{document}

\begin{frontmatter}

\title{The Fellowship of the Dyson Ring: ACT\&Friends' Results and Methods for GTOC~11}


\author[esa]{Marcus M\"artens}
\author[esa]{Dario Izzo}
\author[esa]{Emmanuel Blazquez}
\author[esa]{Moritz von Looz} 
\author[esa]{Pablo G\'{o}mez} 
\author[esa]{Anne Mergy} 

\author[aff_giacomo]{Giacomo Acciarini} 
\author[aff]{Chit Hong Yam} 
\author[aff]{Javier Hernando Ayuso} 
\author[gt]{Yuri Shimane} 

\address[esa]{European Space Agency, Noordwijk, 2201 AZ, The Netherlands}
\address[aff_giacomo]{Department of Computer Science, University of Oxford, Parks Road, OX1 3PJ, Oxford, United Kingdom}
\address[aff]{Mission Design and Operations Group, ispace inc., Sumitomo Fudosan Hamacho Building 3F, 3-42-3, Nihonbashi Hamacho, Chuo-ku, Tokyo, Japan 103-0007}
\address[gt]{School of Aerospace Engineering, Georgia Institute of Technology, Atlanta, Georgia 30332}

\begin{abstract}
Dyson spheres are hypothetical megastructures encircling stars in order to harvest most of their energy output. During the 11th edition of the GTOC challenge, participants were tasked with a complex trajectory planning related to the construction of a precursor Dyson structure, a heliocentric ring made of twelve stations. To this purpose, we developed several new approaches that synthesize techniques from machine learning, combinatorial optimization, planning and scheduling, and evolutionary optimization effectively integrated into a fully automated pipeline. These include a machine learned transfer time estimator, improving the established Edelbaum approximation and thus better informing a Lazy Race Tree Search to identify and collect asteroids with high arrival mass for the stations; a series of optimally-phased low-thrust transfers to all stations computed by indirect optimization techniques, exploiting the synodic periodicity of the system; and a modified Hungarian scheduling algorithm, which utilizes evolutionary techniques to arrange a mass-balanced arrival schedule out of all transfer possibilities. We describe the steps of our pipeline in detail with a special focus on how our approaches mutually benefit from each other. Lastly, we outline and analyze the final solution of our team, ACT\&Friends, which ranked second at the GTOC~11 challenge.
\end{abstract}


\end{frontmatter}


\section{Introduction}

The capability of harvesting energy has been frequently stated as a measurement of the technological advancement of civilizations, most famously by Nikolai Kardashev~\cite{kardashev1964transmission}, after which the Kardashev-scale has been named. A type-II civilization for example is defined as a civilization that is capable of utilizing the complete energy output of its home star, which would be roughly $4 \cdot 10^{26}$ J/s in case of our own sun. A swarm of solar harvesting satellites orbiting the star in dense formation has been proposed as enabling technology for such an ambition. 

While this idea has been explored in science fiction literature before, its first scientific treatment is attributed and named after Freeman Dyson~\cite{dyson1960search}. Dyson's goal was to show the physical plausibility of a giant energy harvesting biosphere by calculations based on our own solar system, leading to important implications with regards to the search for extraterrestrial life. While engineering aspects and astrodynamic specifics were not discussed in Dyson's original work, his thought experiment sparked the imagination of countless minds and served as inspiration for the Global Trajectory Optimization Competition (GTOC) in its 11th edition~\cite{Shen2022}.

The setup of this challenge tasked its participants with the assembly of a precursor structure, a Dyson ring, consisting of twelve equally distributed stations on a circular orbit situated in our solar system. An ensemble of ten mother ships needs to be launched from Earth towards the outer solar system to collect the necessary mass for construction. This collection is performed through the deployment of a multifunctional asteroid transfer device (ATD), which serves as an abstraction for the future technical systems required to effectively mine and transport the resources of such a body. Once an ATD gets activated, part of the asteroid's own mass will be continuously converted into constant acceleration, allowing it to travel towards the construction orbit. Synchronizing the arrival of such asteroids well-phased within a narrow time interval for building while maintaining an optimal arrival mass presents a complex and unique scheduling challenge, including trade-offs between the construction orbit, activation times, mass distribution among stations and combinatorial selection of suitable candidate asteroids. While the actual construction of megastructures at the envisioned scale of the GTOC~11 seems still out of reach, the basic concepts, for example, of asteroid mining~\cite{andrews2015defining, hellgren2016asteroid} and solar powered satellites~\cite{summerer2005advanced, flournoy2011solar} are regarded as emerging space technologies, that will enable us to exploit extraterrestrial resources on an unprecedented scale.
Moreover, the challenging planning and scheduling aspects of large-scale space constructions provide a compelling environment for which multiple different optimization techniques are required.

\begin{figure}[ht]
	\centering
     \includegraphics[width=.9\textwidth]{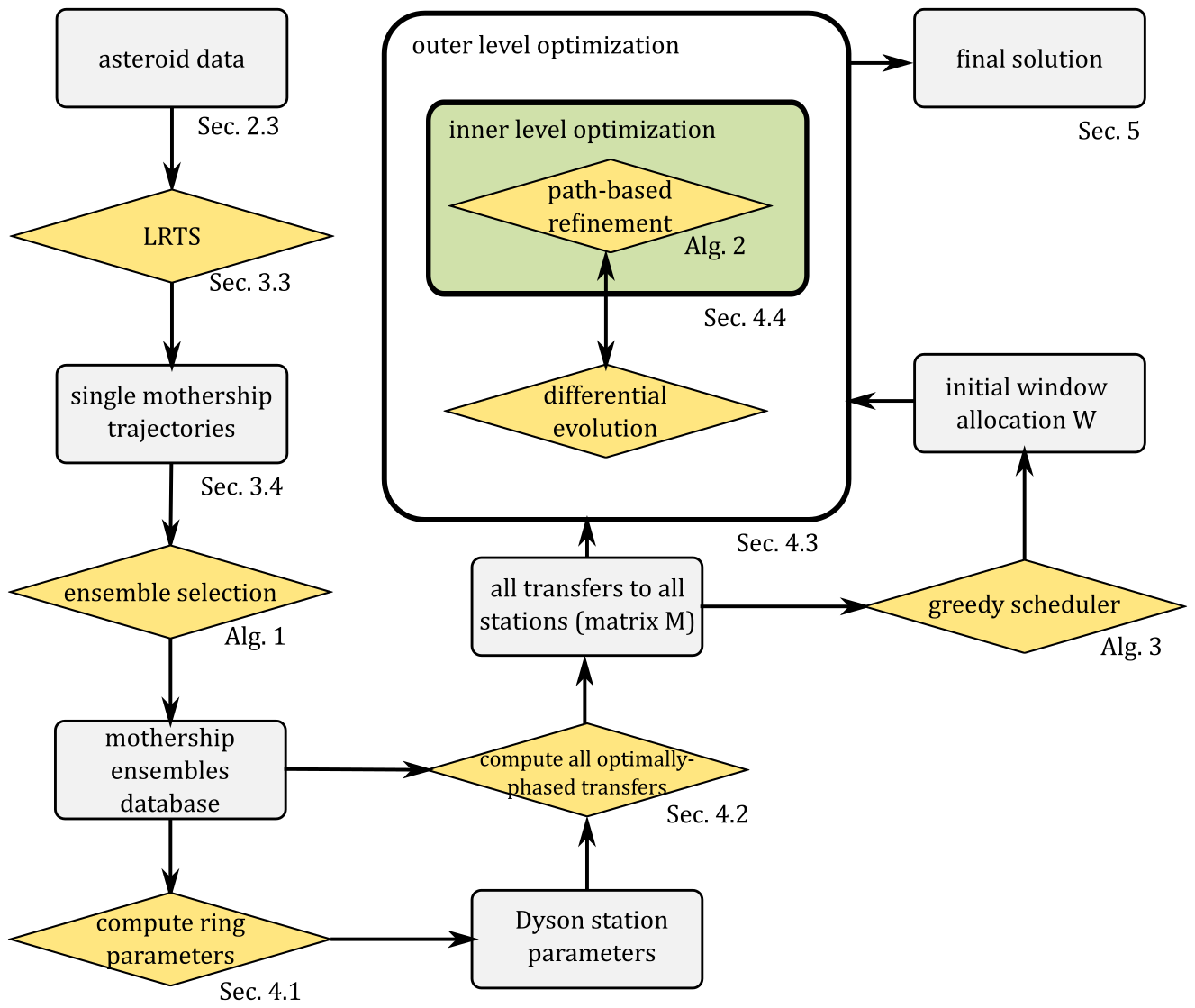}
    \caption{Schematic of solution pipeline. When appropriate, sections and algorithms describing the procedures (yellow diamond) or intermediate results (rectangles) are given.}    
    \label{fig:flowchart}     
\end{figure}

On a high level, any solution strategy to the GTOC~11 challenge needs to address the sub-challenges of 1. asteroid selection, 2. station orbit determination and 3. construction time scheduling. The asteroid selection sub-challenge can be seen as a combinatorial problem that requires the selection of favorable sequences of asteroids linked by optimal transfers. The dominant solution strategies for this type of problem are tree searches, ant colony optimization or hybridisations of both~\cite{simoes2017multi}. An additional difficulty is introduced by the concurrency of the selection, as (ideally) coordination between all ten motherships is exploited. The station orbit determination sub-challenge is strongly tied to the solutions of the other sub-challenges as it links multiple objectives together (as we will show in our preliminary analysis of the GTOC~11 objective function). Tackling it after the asteroid selection decreases its difficulty considerably, as an optimal orbit for a small subset of asteroids can be modelled and solved as a non-linear optimization problem. While this has the downside that the asteroid selection is blind (i.e. it is not informed about the optimal station orbit) an earlier solution of this sub-challenge implies solving the GTOC~11 holistically by directly optimizing for the objective function, which is by design of the competition practically impossible. Lastly, the construction time scheduling sub-challenge is unique and difficult to compare to any established scheduling or sequencing problems that the authors are aware of. In particular, the fact that asteroids and stations move along orbits together with the tight construction constraints demand the development of entirely new algorithms, as it is in the spirit of GTOC. The purpose of this work is to describe the end-to-end optimization pipeline our team used to submit solutions during the competition: a pipeline that integrates scheduling algorithms, evolutionary optimization, machine learning and space flight mechanics routines and solves the three sub-challenges in the aforementioned order. Figure~\ref{fig:flowchart} shows a schematic of this pipeline with references to the corresponding sections and algorithms.

We begin this report with a preliminary analysis of the GTOC~11 objective function, the related continuous-thrust optimal control problem and the population of candidate asteroids provided as part of the challenge setup. Next, we describe how a Lazy Race Tree Search can be designed and improved by a time estimation provided by supervised machine learning, leading to a quick generation of a large set of possible mother ship trajectories and related ensembles. The next section describes our ring building pipeline, including the determination of its orbital parameters, the optimization of the activated asteroid trajectories and the solution of related assignment problems by the Hungarian algorithm in combination with evolutionary optimization techniques. Following these conceptual explanations, our result section describes the generation of our specific submission, which enabled our team, ACT\&Friends, to obtain the second overall rank in the final GTOC~11 leaderboard. We conclude with an outlook on the lessons learned from the competition and point out directions demanding further investigation.

\section{Preliminary Analysis}

\subsection{Objective Function}
The objective function, as released as performance index in the original problem description, has the form:
\begin{equation}
J = B \cdot \frac{10^{-10} \cdot M_{min}}{a^2_{D} \sum\limits_{k=1}^{10}\left(1 + \Delta V_{k}^{Total} / 50 \right)^2}
\label{eq:objfun}    
\end{equation}
where $B$ is a time penalty factor: a number decreasing as the competition advanced and designed as to stimulate early submissions and an active leaderboard.
$M_{min}$ is the minimum cumulative mass delivered at one of the twelve stations. This term encourages to evenly deliver asteroid mass to all target stations and to activate more asteroids.
The target Dyson ring radius $a_D$ appears explicitly only in the denominator, benefitting designs with small ring radii. The Dyson ring radius also factors into the determination of the asteroid mass delivered (numerator) and thus an optimal configuration exists at some unknown value.
Finally, $\Delta V_{k}^{Total}$ is the total velocity increment needed by each mother ship to deliver the ATDs and thus enabling asteroid activation. This term encourages to find fuel efficient trajectories in general, as well as to decrease the number of activated asteroids, assuming that significant velocity changes are required in between asteroid visits. The trade-offs and exact balances of all these contributions can only be made approaching the problem holistically, accounting for the actual asteroid orbits and spacecraft dynamics. Nevertheless, a preliminary qualitative analysis is possible by introducing a few simplifying assumptions.

Let us assume that the final configuration is made by twelve stations, each receiving $N$ identical asteroids with final mass $m_{arr}(a_D)$. The mass delivered depends on the Dyson ring radius, as smaller rings will require more time to be reached resulting in a lower mass at arrival.
We also assume that, on average, the cost for a mother ship to make one transfer and thus activate one asteroid is $\Delta \tilde V$ and is uniform across the trajectories of the ten mother ships. Since the total number of asteroids to be activated is, under these definitions, $12 N$ and they are activated by ten spacecraft, it follows that each mother ship will use a total velocity increment $\Delta V_{k}^{Total} = \frac{12}{10} N \Delta \tilde V$. We can therefore rewrite the objective function, neglecting the various constants, as to highlight all of its trade-offs as:
\begin{equation}
J \propto \frac{N m_{arr}(a_D)}{a^2_{D} \left(1 + \frac{12}{10} N \Delta \tilde V / 50 \right)^2}
\label{eq:objfun_simple}    
\end{equation}
Using the above equation and setting its derivative with respect to $N$ to zero, we find that:
$$
\frac{\partial J}{\partial N} =0 \rightarrow N^* = \frac{41.6}{\Delta \tilde V}
$$
which tells us that spending, for example, $1$ km/s on average to activate one asteroid, should result in each mother ship activating around 42 asteroids. Note that no team during the competition submitted solutions quantitatively in this regime, suggesting the impossibility to fulfill the hypothesis made here on the Dyson ring's final configuration, but maybe also indicating the possibility to further improve the trajectories found so far.
The expression above is interesting, nevertheless, as it states that the optimal number of asteroids delivered to each station does not depend on the chosen Dyson ring radius, but only on the fuel efficiency of the mother ship trajectories. It also reveals the various trade-offs clearly, showing how more activated asteroids (i.e. higher $N$) do not necessarily result in a better score if the cost paid by the mother ship to activate them is too high. 
As for the Dyson ring radius, the expression also highlights how its optimal value depends exclusively on the trade-off dictated by the quantity $\frac {m_{arr}(a_D)}{a^2_{D}}$. By introducing additional assumptions, a more detailed analysis could also take the derivative with respect to $a_D$ into account, approximating the functional form for $m_{arr}(a_D)$, for example, making use of the Edelbaum approximation \cite{edelbaum1961propulsion}. However, in this report we will not follow this idea and move on by introducing the optimal control problem associated with the asteroid transfers.

\subsection{Low Thrust Continuous Transfers}
\label{subsec:low_thrust_transfers}
The trajectory of an asteroid after its activation is determined by a constant acceleration $\Gamma = 10^{-4} m/s^2$ acting along the direction $\hat{\mathbf i}(t)$.
The associated time-optimal control problem (to find $\hat{\mathbf i}(t)$) can be efficiently solved by either direct or indirect methods~\cite{von1992direct}.
After experimenting with both approaches, we decided to use indirect methods exclusively as they offered several advantages, e.g. their numerical precision was most of the time compliant with the error tolerances of $10$ km in position and $0.01$ m/s demanded by the GTOC~11 problem statement.
Applying the approach from Pontryagin's theory~\cite{pontryagin1962maximum}, we started from the simple equations of motion in Cartesian coordinates: 
\begin{equation}
\label{eq:dyn}
\left\{ 
\begin{array}{l}
    \dot{\mathbf{r}} = \mathbf{v} \\
    \dot{\mathbf{v}} = -\frac{\mu}{r^3}\mathbf{r} +\Gamma\mathbf{\hat{i}}(t)
\end{array}
\right.
\end{equation}
where $\mu$ is the gravitational constant of the Sun. We thus considered the problem of finding $t_f$, $t_f-t_0$ and a function $\mathbf{\hat{i}}(t)$ with $t \in [t_0, t_f]$ so that, under the dynamics defined in Eq.(\ref{eq:dyn}), the state is steered from the initial state $\mathbf r_{ast}(t_0), \mathbf v_{ast}(t_0)$ representing some asteroid ephemerides in $t_0$ to the target state $\mathbf r_{st}(t_f),\mathbf v_{st}(t_f)$ representing the target station position and velocity. 
The following cost function needs to be minimal:
$J = t_f-t_0 = \int^{t_f}_{t_0} 1 dt$, leading to the introduction of the Hamiltonian: 

\begin{equation}
\label{eq:hamiltonian}
    \mathcal H(\mathbf{r},\mathbf{v},\boldsymbol{\lambda}_{\mathbf r},\boldsymbol{\lambda}_{\mathbf v},\mathbf{\hat{i}}) = \boldsymbol{\lambda}_{\mathbf r}\cdot \mathbf{v}+\boldsymbol{\lambda}_{\mathbf v}\cdot\bigg(-\frac{\mu}{r^3}\mathbf{r}+\Gamma\mathbf{\hat{i}}\bigg)+1
\end{equation}
where we have introduced the co-state's functions $\boldsymbol{\lambda}_{\mathbf r}$ and $\boldsymbol{\lambda}_{\mathbf v}$. It is therefore straightforward to obtain the following consequence of the Pontryagin's maximum principle:

\begin{equation}
\label{eq:duh}
\mathbf{\hat{i}}=-\frac{ \boldsymbol{\lambda}_{\mathbf v}}{\lambda_v}
\end{equation}
which defines the optimal direction of the acceleration, and the corresponding augmented dynamics derived from the Hamiltonian performing the derivatives $\dot {\mathbf x} = \frac{\partial \mathcal H}{\partial \boldsymbol \lambda}$, $\dot {\boldsymbol \lambda} = - \frac{\partial \mathcal H}{\partial \mathbf x}$:
\begin{equation}
\label{eq:augdyn}
\left\{ 
\begin{array}{l}
    \dot{\mathbf{r}} = \mathbf{v} \\
    \dot{\mathbf{v}} = -\frac{\mu}{r^3}\mathbf{r}-\Gamma\frac{ \boldsymbol{\lambda}_{\mathbf v}}{\lambda_v}\\
    \dot{\boldsymbol\lambda}_{\mathbf{r}} = \mu \left(\frac{\boldsymbol{\lambda}_{\boldsymbol v}}{r^3} - 3(\boldsymbol \lambda_{\boldsymbol v}\cdot\mathbf r)\frac{\mathbf r}{r^5} \right)  \\
    \dot{\boldsymbol\lambda}_{\mathbf{v}} = - \boldsymbol{\lambda}_{\mathbf r}
\end{array}
\right.
\end{equation}
According to Pontryagin's theory, an optimal transfer will necessarily be a solution to the above differential equations with the added condition $\mathcal H=0$ as we are also considering a free time problem. 
Note however that this last condition is not necessary since we can always multiply the co-states for some coefficient $\lambda_0$ and get a new equivalent solution resulting in $\mathcal H=0$ since all the relevant equations are homogeneous in the co-states. 
We will nevertheless seek also a zero Hamiltonian solution as to avoid numerical instabilities. 
Consider now the shooting function:

\begin{equation}
    \Phi_{t_f}(t_0, \boldsymbol{\lambda}_{\mathbf{r0}},  \boldsymbol{\lambda}_{\mathbf{v0}}) = [\mathbf r(t_0+T) - \mathbf r_{st}(t_f), \mathbf v(t_0+T) - \mathbf v_{st}(t_f), \mathcal H]
\end{equation}

For any given $t_f$, assuming $t_f = t_0+T$, the roots of the above function allow us to find the optimal transfer for an asteroid. Clearly, if the resulting $t_0$ is before the activation date, there is no valid solution arriving in $t_f$ at the chosen station. Formally, when looking for optimal asteroid transfers to a given station, we thus solve the following rendezvous optimization problem:

\begin{equation}
\label{eq:rdvz}
    \begin{array}{rl}
        \mbox{find:} &  t_f, t_0, \boldsymbol{\lambda}_{\mathbf{r0}},  \boldsymbol{\lambda}_{\mathbf{v0}}\\
        \mbox{to minimize:} & t_f-t_0\\
        \mbox{subject to:}&  t_f = T + t_0 \\
        &  \Phi_{t_f}(t_0, \boldsymbol{\lambda}_{\mathbf{r0}},  \boldsymbol{\lambda}_{\mathbf{v0}}) = \mathbf 0 \\
        & t_0 > t^{act} \\
        & t_f \in [\underline t_f, \overline t_f]
    \end{array}
\end{equation}

where an arrival window $[\underline t_f, \overline t_f]$ is assumed, and $t^{act}$ indicates the asteroid activation epoch.
Similarly, when considering a phase-free problem (for example for the Edelbaum approximation learning), we considered the following version where the time of flight $T$ is no longer linked to the starting and arrival position (a simple numerical trick to get the phase-free version of the same problem):

\begin{equation}
\label{eq:phasefree}
    \begin{array}{rl}
        \mbox{find:} &  T, t_f, t_0, \boldsymbol{\lambda}_{\mathbf{r0}},  \boldsymbol{\lambda}_{\mathbf{v0}}\\
        \mbox{to minimize:} & T\\
        \mbox{subject to:}
        &  \Phi_{t_f}(t_0, \boldsymbol{\lambda}_{\mathbf{r0}},  \boldsymbol{\lambda}_{\mathbf{v0}}) = \mathbf 0 
    \end{array}
\end{equation}

It is worth mentioning that we experimented also in writing the shooting function in terms of equinoctial elements instead of Cartesian, observing an improvement in the convergence of the sequential quadratic programming method we employed to solve the problems. 
Although comparing computational times for numerically solving specific classes of problems is arguably involved due to the dependencies related to the deployed hardware and software on top of the problem specific conditions and parameters, a rough indication on the order of magnitude can be made. One instance of either problem defined above is solvable on a modern single threaded CPU in one second on average. Note that the guess provided on the initial value of each of the co-states is here a uniform random number in the bounds [-10,10].
Some local minima are present, though, and therefore multiple starts may provide better solutions.
This relatively short computational time allowed us to consider strategies where solving large numbers of such problem instances is obligatory.
On a more technical note, we used a Taylor-based method (Heyoka~\cite{biscani2021revisiting}) for all numerical propagation of the augmented dynamical equations and SNOPT~\cite{gill2005snopt} as implementation of the SQP method for solving the optimization problems transcribed as a non-linear program (NLP) using the pagmo/pygmo package~\cite{biscani2020parallel}.

\subsection{Asteroids}
The core data for the competition problem lies in the provided dataset of $83\,453$ asteroids, defining the principal sources of mass available for transport to the twelve ring stations. In addition to the orbital elements specifying the asteroids' orbits, each asteroid has an initial mass $m_0$ between $2.266 \cdot 10^{9}$ and $m_{max} = 1.963 \cdot 10^{14}$ kg. During its constant acceleration transfer to the target ring, the mass of an asteroid decreases proportionally to its initial mass by a factor of $\alpha = 6 \cdot 10^{-9}$. Consequently, the longer the asteroid needs to reach its final destination, the less of its initial mass will remain for ring construction. Given a transfer time of $T$ and target semi-mayor axis $a_D$, we define the arrival mass $m_{arr}$ of an asteroid as
\begin{equation}
\label{eq:arr_mass}
m_{arr}(a_D) = m_0 \cdot (1 - \alpha \cdot T).
\end{equation}

\begin{figure}[ht]
	\centering
     \includegraphics[width=.9\textwidth]{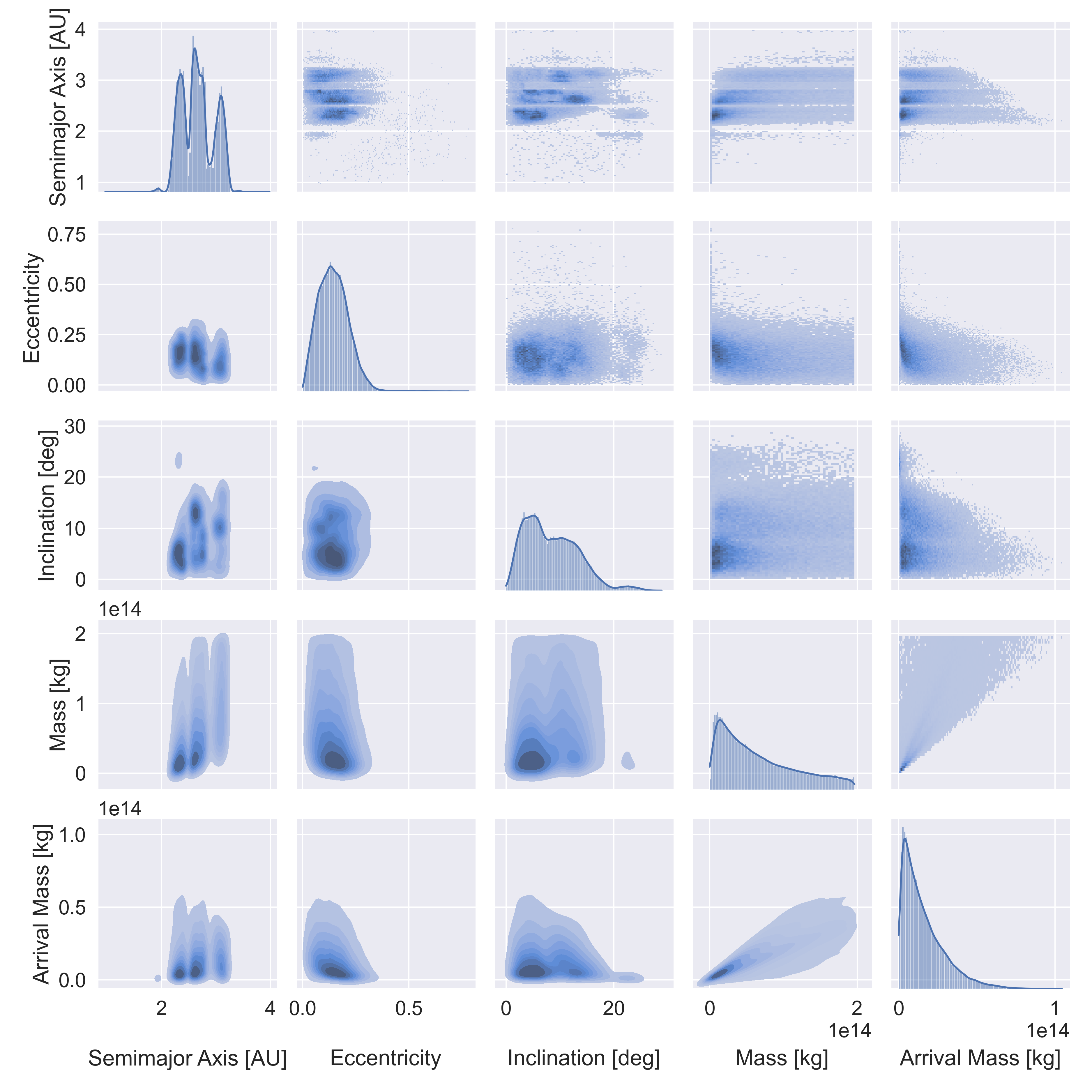}
    \caption{Detailed relationships between critical orbital elements, (initial) mass and arrival mass at the optimal station assuming a Dyson radius of $1.29$ AU at $0^\circ $ inclination. Only asteroids arriving with at least $10^9$ kg are shown ($N = 79,266$).}    
    \label{fig:asteroid_population}     
\end{figure}

Thus, initial mass and orbital elements are crucial for the asteroid selection. Solving the optimal control problem (OCP) as given by Eq.(\ref{eq:phasefree}) for all asteroids to a hypothetical Dyson ring at $1.29$ AU and $0^\circ$ inclination gives a first indication of the mass we can expect to arrive at this particular Dyson ring orbit, shown in Figure~\ref{fig:asteroid_population} for all asteroids that arrive with at least $10^9$ kg ($N = 79,266$). Notably, the semi-major axis of these asteroids features a trimodal distribution with peaks at $\approx 2.33$, $2.67$ and $3.15$ AU. Asteroids at larger distances lead to a reduced arrival mass given their prolonged transfer times.
Eccentricity and inclination are on average limited for most asteroids ( $0.150 \pm 0.0783 $ and $ 9.21^\circ \pm 6.14^\circ$, respectively) with higher eccentricities and inclinations also leading to a lower arrival mass. The latter, however, is a slightly biased result given our assumption of the $0^\circ$ inclination of the target Dyson ring.
The initial mass distribution seems to be independent of semi-major axis, inclination and eccentricity, with an average of $6.71 \pm 5.21 \cdot 10^{13}$ kg. The majority of asteroids feature a smaller mass with only $25.9$\% having a mass greater than $10^{14}$ kg. The arrival mass is on average $1.52 \pm 1.38 \cdot 10^{13}$ kg and shows a clear relationship to all other displayed parameters and its distribution has a higher kurtosis compared to the initial mass distribution. Only $2.8$\% of all asteroids have an arrival mass larger than $5 \cdot 10^{13}$ kg.

In summary, we deduce the following insights from our analysis: First, having a method to reliably estimate transfer time $T$ and thus arrival mass $m_{arr}(a_D)$ is essential for any high level strategy of asteroid selection. Second, the dataset of asteroids can be reduced considerably if focus is given to the long tail of asteroids providing higher arrival masses. Finally, constructing a Dyson ring of small radius (e.g. close to the minimum possible of $0.65$ AU) results in a relatively poor payoff, as already at $1.29$ AU further out asteroids bring in comparatively little mass.

\section{Mother Ship Trajectory Design}

\subsection{Earth-to-Asteroid and Asteroid-to-Asteroid Transfers}
\label{subsec:a2a}

The task of each mother ship is to depart from Earth and deploy a number of ATDs (Asteroid Transfer Devices) within a limited mission time of $20$ years. As such, the Earth-to-Asteroid (E2A) and the Asteroid-to-Asteroid (A2A) transfers are basic building blocks for the assembly of a trajectory. While the problem statement allows for up to four impulsive velocity changes between transfers, our search strategy is based on the rapid optimization of three-impulse legs, containing a departure, deep space and arrival impulse. The three impulse strategy was deemed, at this point, to be a good compromise between computational complexity and optimality of the resulting transfers. In hindsight, since most optimal transfers requested anyway only two impulses, the whole mother ship design could have been simplified considerably.

For any given target asteroid, the E2A optimization problem is defined as follows:

\begin{equation}
    \begin{array}{rl}
        \mbox{find:} &  t_0, T, V_{\inf}, u, v, \eta\\
        \mbox{to minimize:} & \Delta V\\
    \end{array}
\end{equation}
where $t_0$ is the starting epoch, $T$ the transfer time, $ V_{\inf}$ is the relative spacecraft velocity at departure and $u$ and $v$ are angles indicating the direction of departure. Following the competition setup, $V_{\inf}$ is discounted by $6$km/s but can be higher to allow for faster transfer times. After this departure impulse, the mother ship is propagated for $\eta T$ towards its initial direction until another impulse can be applied (deep space maneuver). From there, a Lambert arc connecting to the target asteroid is constructed and the final impulse is adjusted to match the asteroid's velocity up to $2$km/s for the deployment of the ATD.

Following the deployment of the first ATD, the starting epoch, position and initial velocity of the spacecraft is fixed for the following transfer. Thus, the A2A optimization problem has one less decision variable, but is otherwise identical:

\begin{equation}
\label{eq:a2a}
    \begin{array}{rl}
        \mbox{find:} &   T, V_{\inf}, u, v, \eta\\
        \mbox{to minimize:} & \Delta V\\
    \end{array}
\end{equation}

During our tree searches, we solve millions of the above defined optimization problems on the fly with jDE~\cite{brest2006self}, a self-adaptive differential evolution algorithm.

The GTOC~11 problem statement does not give any explicit constraints on the maximum $\Delta V$ one mother ship can provide, but the objective function punishes excessive use of it. Thus, the typical trade-off between total transfer time and $\Delta V$ needs to be taken into account. For the basic transfers, our focus is solely on minimizing $\Delta V$ (i.e. the sum of all impulsive velocity changes over the three possible impulses) while keeping an upper bound (for our submitted solution of $380$ days) on $T$. As we show later on, our Lazy Race Tree Search is designed to assemble high ranking long chains of asteroid transfers and thus minimizes for the sum of all $T$ accordingly.

\subsection{Improving the Edelbaum Approximation}
\label{subsec:edelbaum_learn}

\begin{figure}[ht]
	\centering
     \includegraphics[width=.9\textwidth]{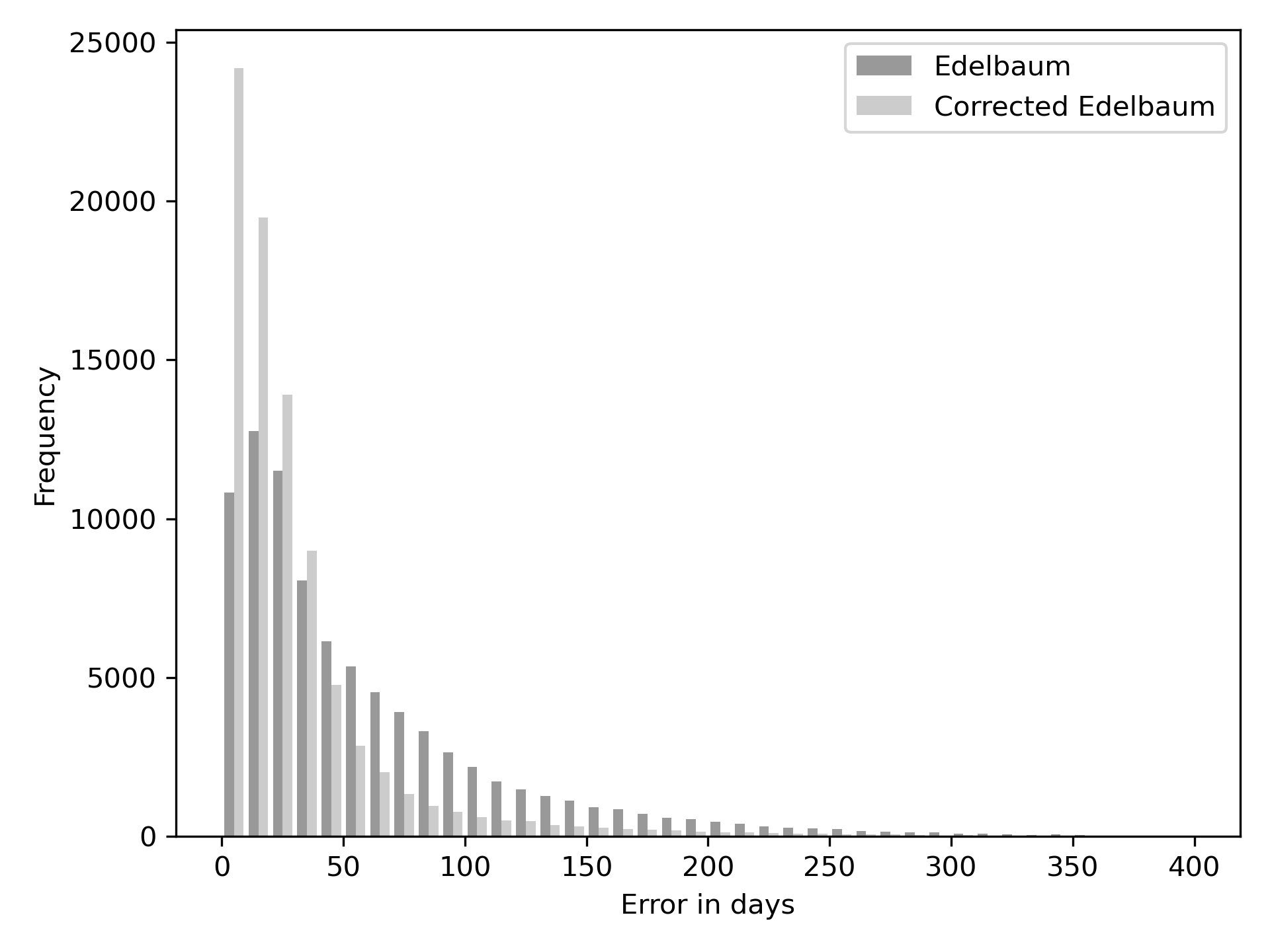}
    \caption{Error distribution across the test set for the Edelbaum approximation (average $\approx 60$ days) and the learned corrected Edelbaum approximation (average error is  $\approx 30$  days)}    
    \label{fig:corrected_edelbaum}     
\end{figure}

Eq.(\ref{eq:arr_mass}) shows that the arrival mass of an asteroid depends largely on its transfer time $T$. A rough estimate $\hat{T}$ of $T$ can be obtained by using an Edelbaum approximation for low-thrust time-optimal trajectories if we assume circular orbits~\cite{kluever2011using}:

\begin{align}
    \label{eq:edeltree}
    \Delta V_{total} &= \sqrt{V_{ast}^2 + V_{st}^2 - 2V_{ast}V_{st} \cos{\left[(\pi/2)\Delta i\right]}}\\
    \hat{T} &= \frac{\Delta V_{total}}{\Gamma}
\end{align}

where $V_{\text{ast}} = \sqrt{\mu / a_{ast}}, V_{st} = \sqrt{\mu / a_{st}}$ are the circular velocities at departure and arrival, $\Delta i = |i_{st} - i_{ast}|$ the desired inclination change and $\Gamma$ the low-thrust acceleration.
While $\hat{T}$ is fast to compute, it neither accounts for the eccentricity nor the argument of perigee of the departure orbit, resulting in an error $T^* - \hat{T}$ with regards to the optimal $T^*$ of the phase-free transfer as given by the solution of the corresponding OCP as given by Eq.(\ref{eq:phasefree}). The fact that a sufficiently large population of such problems can be constructed and solved independently in parallel, allows us to apply supervised machine learning.

In particular, we are interested in regressing the error of the Edelbaum transfer time $\hat{T}$:
\begin{equation}
    T^* - \hat{T} = f(a_{st}, a_{ast}, i_{ast}, e_{ast}, \omega_{ast})
\end{equation}
For this purpose, we generated a database of $375\,000$ solved optimal control problems by sampling $a_{st}$ uniformly at random within $[0.9, 1.4]$ AU and selecting asteroids randomly from all $83\,453$ available candidates.
The values used to bound $a_{st}$ try to bracket as closely as possible our intuition on its optimal value and were not chosen following a rigorous approach. 
$250\,000$ instances were used to train a multi-layer perceptron of five hidden layers of $50$ neurons with ReLU activation functions. 
The remaining instances were used to evaluate the performance of the corrected Edelbaum approximation. 
Figure~\ref{fig:corrected_edelbaum} shows the comparison between the uncorrected and our corrected Edelbaum approximation. As the decrease in error allows for more accurate predictions, we apply this improved approximator (denoted by $T_e$ in the following) to our tree search instead of the original Edelbaum estimation, reducing its error in almost all cases. 
The impact on the run time of the tree search is negligible, as the inference of the neural network (once loaded into memory) is fast.

\subsection{Lazy Race Tree Search}

The Lazy Race Tree Search (LRTS) is a high level search strategy addressing combinatorial decisions for trajectory optimization problems. It is best understood as a variant of the widely known beam search~\cite{bisiani1992beam, wilt2010comparison}, which itself is a generalization of the basic greedy search paradigm. In a greedy search, one would grow a chain of asteroids one by one by selecting only the best asteroid (according to some ranking criterion) at each step. However, for the asteroid selection problem, an optimal selection within each step does not imply optimality of the complete chain, i.e. Bellman's principle of optimality cannot be applied. Consequently, a greedy search will almost always be suboptimal.

Beam search improves on the greedy search by considering not only the single best solution, but by keeping the $g$ best next asteroids saved inside a search tree structure, where $g$ is sometimes called ``branch-factor'' or ``fanout''. If one would allow this search tree to simply keep on branching, a chain of length $k$ would grow exponentially (i.e. $k^b$) which is why the beam search additionally limits the number of partial solution at each level of the search tree by the beam-width parameter $b$. In other words, each level of the tree prunes all intermediate solutions down to the $b$ best. This means that the search tree of a beam-search for a chain of $k$ asteroids requires at most $k \cdot b \cdot g$ evaluations, making it efficient to store and compute in practice while not being overly greedy.

The LRTS improves on the beam search concept by defining fair ranking criteria that lead themselves naturally to problems where ongoing time is an important aspect. It was developed and deployed successfully already during the GTOC 6, which required the assembly of a long multi-leg flyby tour around the Galilean moons~\cite{Izzo2013Search}. Similarly to this challenge, our goal is the assembly of long asteroid chains with high arrival mass while maintaining a low $\Delta V$. To achieve this goal, LRTS incrementally builds a search tree by expanding and pruning of nodes until the total mission time is exhausted. Each node in the search tree constitutes a chain of transfers beginning from Earth and visiting a number of asteroids $a_1, \ldots, a_k$.

The specialty of LRTS is that it is not restricted to comparing solutions based on their chain length like a traditional beam search would do. Instead, nodes are ranked within a number of $s$ time-slices, beginning with the earliest time-slices (i.e. with the nodes that are ``slowest'' in the current race). The ranking of each node is computed by a variation of the objective function given in Eq.(\ref{eq:objfun}) that ignores the bonus component and assumes ten identical mother ships:

\begin{equation}
\label{eq:lrts_ranking}    
J'(a_1, \ldots, a_k) = \frac{  M_e }{a^2_{D} \cdot \left(1 + \Delta V_{k}^{Total} / 50 \right)^2}
\end{equation}
where $M_e$ is the sum of all expected arrival masses from all collected asteroids assuming a transfer time computed by the improved Edelbaum approximator. If the number of nodes within a time slice exceeds a certain beam width parameter $b$, the lowest ranking nodes are pruned. Thus, only the $b$ highest ranked nodes are expanded further into the future.

\begin{figure}[ht]
	\centering
     \includegraphics[width=.9\textwidth]{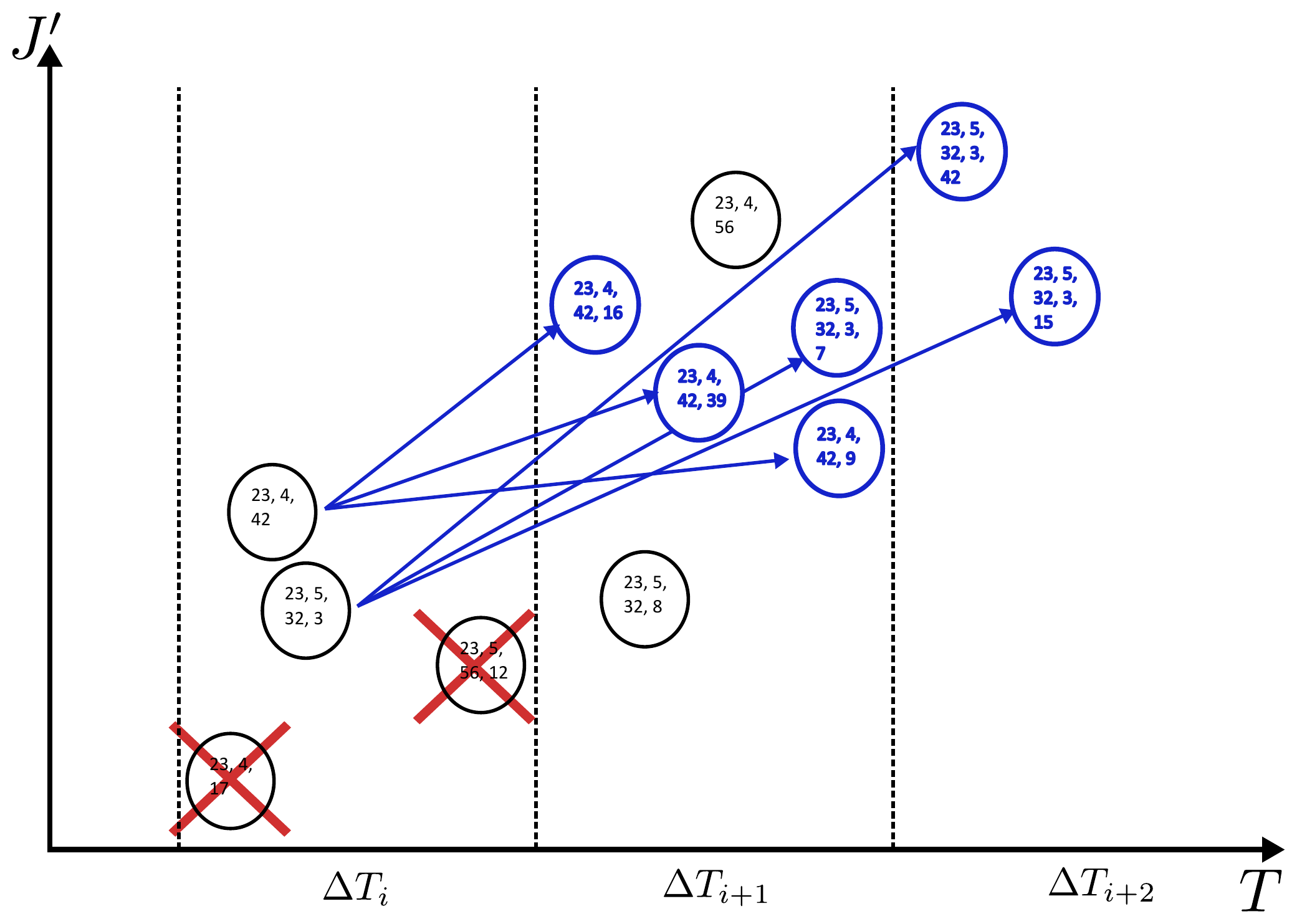}
    \caption{LRTS working on time-slice $\Delta T_i$. The lower ranked nodes are pruned, the $b = 2$ highest ranked nodes are each expanded in $g = 3$ new nodes (bold and blue), sorted into different time-slices. The numbers inside the nodes correspond to the visited asteroid sequence. Note that LRTS ranks sequences of different length within the same time-slice.}    
    \label{fig:lrts}     
\end{figure}

Given a node with asteroid sequence $a_1, \ldots, a_k$, the expansion (or branching) step of a node expands the sequence by another asteroid $a_{k+1}$. For that purpose, the final conditions of arrival at $a_k$ are considered and the orbital phasing indicator~\cite{Izzo2016Designing} is computed to determine a distance from $a_k$ to all unvisited candidate asteroids. For a given fan-out parameter $g$, a kd-tree~\cite{maneewongvatana1999s} is constructed to preselect the $2g$ highest ranking asteroids according to this indicator. For each preselected asteroid candidate, we then construct and solve an A2A-optimzation problem (see Eq.(\ref{eq:a2a})) to determine the necessary $\Delta V$ and transfer time (asteroid to asteroid) for this leg. Additionally, we use the improved Edelbaum approximator to estimate the transfer time $T_e$ of the asteroid to the ring station for a pre-defined semi-mayor axis $a_{D}$, resulting in an estimate of its arrival mass. This allows us to rank all $2g$ candidates according to their improvement on $J'$. Subsequently, only the best $g$ new nodes are inserted in their corresponding time-slice of the search tree and the rest is discarded.
Thus, expansion and pruning are both guided by $J'$, which takes all important factors of the original objective function into account. The lazy race aspect of ranking on time-slices assures a fair comparison which is unbiased by the length $k$ of the corresponding asteroids chains, which is not necessarily correlated with the objective function. Figure~\ref{fig:lrts} shows conceptually how LRTS operates for one slice of time.
LRTS terminates if there is no node in the search tree left that can be expanded, as the transfer time to the next asteroid would be outside the mission time. After termination, the highest ranking nodes according to $J'$ are extracted from the search tree, each of them describing a valid and complete mother ship trajectory. 

\subsection{Mother Ship Ensemble Selection}
\label{subsec:mothership_ensemble_selection}

\begin{algorithm}[ht!]
\caption{Mother ship ensemble selection heuristic}
\label{alg:greedy}
\begin{algorithmic}
\Require $J_{init}, \delta, \mathcal{U} = \{S_1, S_2, S_3, \ldots \}$

\State $J_{\tau} \gets J_{init}$
\State $\mathcal{S} = \emptyset$
\While{$|\mathcal{S}| < 10$}
    \State $P \gets \{S \in \mathcal{U} \,\backslash\, \mathcal{S} ~|~ J'(S) > J_{\tau}\}$
    \State $S_R \gets$ uniformly at random from $P$
\If{$S_i \cap S_R = \emptyset$ for all $S_i \in \mathcal{S}$}
    \State $\mathcal{S} \gets \mathcal{S} \cup \{S_R\}$
\ElsIf{$|\bigcup\limits_{S \in \mathcal{S}} S \cap S_R| = 1$}
    \State $a \gets \bigcup\limits_{S \in \mathcal{S}} S \cap S_R$ \Comment{$a$ is a single asteroid}
    \State $S_R' \gets S_R \,\backslash\, \{ a \}$
    \State $\mathcal{S}' \gets \mathcal{S}$ removing $a$ from conflicting trajectory in $\mathcal{S}$
    \If{$\sum\limits_{S \in \mathcal{S} \cup \{ S_R' \}}\frac{J'(S)}{|\mathcal{S} + 1|} > J_{tr}$ or $\sum\limits_{S \in \mathcal{S'} \cup \{ S_R \}}\frac{J'(S)}{|\mathcal{S} + 1|} > J_{tr}$}
        \If{$\sum\limits_{S \in \mathcal{S} \cup \{ S_R' \}}\frac{J'(S)}{|\mathcal{S} + 1|} > \sum\limits_{S \in \mathcal{S'} \cup \{ S_R \}}\frac{J'(S)}{|\mathcal{S} + 1|}$}
            \State $\mathcal{S} \gets \mathcal{S} \cup \{S_R'\}$
        \Else
            \State $\mathcal{S} \gets \mathcal{S'} \cup \{S_R\}$
        \EndIf
    \EndIf
\Else
    \State $J_{\tau} \gets J_{\tau} - \delta$  \Comment{$\delta$ is a small constant}
\EndIf
\EndWhile
\end{algorithmic}
\end{algorithm}

Considering a large amount of possible mother ship trajectories and their corresponding sequence of visited asteroid, the likelihood of having the same asteroid in two or more of these sequences is naturally increasing. If one would select two or more of such sequences visiting the same asteroid at different times, it would violate the validity of the solution according to the problem statement. In fact, the asteroid in question might not be even at the same position when the second visit occurs, as it could already have started moving towards the stations and thus altering its orbit. To circumvent this issue, our ensemble selection is built incrementally by only adding asteroid sequences that are pairwise disjoint to previously selected sequences or that can be fixed (by removing a double visited asteroid from one conflicting sequence) without sacrificing too much of (expected) $J$.

Formalizing this idea, we model every single mother ship trajectory as a set $S_i = \{a_1, \ldots, a_k\}$ of asteroids. The problem is now to find a collection of 10 such sets $\mathcal{S} = \{S_1, \ldots, S_{10}\}$, to which we refer to as mother ship ensemble, such that the expected $J(\mathcal{S})$ is maximal while $\mathcal{S}$ is pairwise disjoint, meaning $S_i \cap S_j = \emptyset$ for all $i,~j~\in~\{1,\ldots,10\}$. Assuming a large set $\mathcal{U}$ of many possible trajectories $S_i$, this problem becomes quickly intractable, as it is closely related to the NP-complete maximum independent set problem in graphs. More precisely, if we model all $S_i$ as nodes in a graph and connect all $S_i, S_j$ for which $|S_i \cap S_j| > 0$ holds, the problem is equivalent to finding a $10$-independent set with maximum $J$.

Consequently, we decided to find our mother ship ensemble by a greedy heuristic, which incrementally builds $\mathcal{S}$ by sampling and combining random $S_i \in \mathcal{U}$. Since our goal is to maximize the expected $J$ of the whole ensemble, we bias our sampling on a slowly decreasing threshold value $J_{\tau}$ which limits possible candidates at the early phases of the algorithm to comparatively high $J'$ (oversampling). Furthermore, if the current ensemble $\mathcal{S}$ and $S_i$ only overlap in one asteroid $a$, we check whether removing $a$ from either $S_i$ or the corresponding $S_j \in \mathcal{S}$ would be feasible, and if so, whether the average expected $J'$ from this fixed ensemble would be above $J_{\tau}$, in which case we include $S_i$ into $\mathcal{S}$ and proceed. Noticeably, this type of single-asteroid removal is a feature of our heuristic which goes beyond the above theoretical independent set problem formulation, but also requires some non-negligible time for reoptimization of trajectories. Algorithm~\ref{alg:greedy} describes our heuristic in pseudocode.

\section{Ring Building Pipeline}

Following the construction of ten mother ship trajectories (a mother ship ensemble) in the previous section, the starting point of our ring building pipeline is a corresponding set of $N_{tot}$ asteroids $\mathcal{A} = \bigcup\limits_{S \in \mathcal{S}} S$ and their earliest possible activation epochs $t^{act}_{i}, i = 1 .. N_{tot}$. The pipeline proceeds through three steps: the computation of optimal ring parameters, the optimization of the phased asteroid transfers and the scheduling of their arrival to the target stations. The results of these steps are, correspondingly, the exact orbits of the twelve Dyson ring stations, optimally-phased transfer trajectories for each asteroids and a set of construction time windows for each of the twelve stations, resulting in a complete solution for the GTOC~11 challenge.

\subsection{Computing the Ring Parameters}
\label{subsec:ring_parameters}

The competition requires the orbit of the Dyson ring to be circular $e_D = 0$, but leaves its semi-major axis (and radius) $a_D$, inclination $i_D$ and right ascension of the ascending node $\Omega_D$ free (as well as a starting station phase $\varphi_D$ which we decided to set and keep to zero). Following the objective function Eq.(\ref{eq:objfun}), our design goal is to select these free parameters to allow favorable (i.e. short and well-phased) transfers for all activated asteroids to increase $M_{min}$ while also reducing $a_D$ as much as possible. A precise computation of $M_{min}$ (and thus a direct optimization for $J$) is prohibitive at this point, as we would need to know the exact transfer times for each asteroid and a (potentially optimized) schedule for each station. However, this information becomes only accessible to us at the end of the pipeline and not at its beginning. Consequently, we define an approximate objective function $\bar{J}$ under the following simplifying assumptions:

\begin{itemize}
    \item Instead of the actual phased transfer times of each asteroid in $\mathcal{A}$ to any given station of the Dyson ring, we use the corrected Edelbaum approximation described in Section~\ref{subsec:edelbaum_learn} as a phase-less substitute, denoted by $T_{e,i}$ in the following. 
    \item Only asteroids in $\mathcal{A}$ capable to deliver a non-zero arrival mass to any station before the final epoch $t_f$ are relevant for the ring construction. Thus, we define a subset $\mathcal{A}_e \subset \mathcal{A}$ of relevant asteroids as followed:
\[
    \mathcal{A}_e = \lbrace i \in \mathcal{A} \mid t^{act}_{i} + T_{e,i} < t_f \text{ and } m_{0,i} \cdot (1 - \alpha \cdot T_{e,i}) > 0 \rbrace.
\]
    \item Although $\mathcal{A}_e$ is not necessarily equal to $\mathcal{A}$, we still assume the complete $\Delta V$ required for visiting all asteroids $\mathcal{A}$. Thus, the potential decrease in $\Delta V$ by re-optimizing the mother ship trajectories to only visit the relevant asteroids $\mathcal{A}_e$ is not accounted for at this point.
    \item Imbalances in the mass-distribution of the stations are also not accounted for. Instead, we assume that the maximum deliverable mass
\[
     M_{tot} = \sum\limits_{i \in \mathcal{A}_e}{}m_{0,i} (1- \alpha \cdot T_{e,i}) 
\]    
    is achieved and equally distributed so that each station receives a mass of $M_{tot} / 12$.
\end{itemize}

Given above assumptions, we define the following optimization problem:

\begin{equation}
\label{eq:ringparameters} 
    \begin{array}{rl}
        \mbox{find:} & a_D, i_D, \Omega_D\\
        \mbox{to maximize:} & \bar{J} = \dfrac{\sum\limits_{i \in \mathcal{A}_e}{}m_{0,i} (1- \alpha \cdot T_{e,i})}{a^2_{D} \sum\limits_{k=1}^{10}\left(1 + \Delta V_{k}^{Total} / 50 \right)^2}\\
        \mbox{subject to:}&  0.65 \leq a_D \leq 5 \\
        &  0 \leq i_D \leq \pi \\
        & 0 \leq \Omega_D \leq 2\pi
    \end{array}
\end{equation}

During the competition, we constructed and solved multiple of these optimization problems by Co-variance Matrix Evolutionary Strategy (CMA-ES)~\cite{hansen2001completely}, allowing us to fix the orbital parameters of the ring for the subsequent steps of our pipeline.

\subsection{Optimally-phased Trajectories}
\label{subsec:optimally_phased_traj}
Given the orbital parameters $a_D, i_D$ and $\Omega_D$ of the Dyson ring stations from the previous pipeline step, we now seek to find efficient transfer opportunities for all asteroids $\mathcal{A}$ to all of the twelve target stations. These are trajectories that reach a station time-optimally when the arrival epoch is left free. In this sense, we refer to them as optimally-phased trajectories. Every asteroid should preferably travel along one of these trajectories to its assigned station within the corresponding station building window, as this will ensure an efficient use of its mass. Since the station building windows will be assigned later, we assume for each asteroid $i \in \mathcal{A}$ an arrival interval of $[t_i^{act}, t_f]$, where $t_f$ denotes the end of the whole mission. Thus, given this interval, the task is to find all the local minima of the optimization problem stated in Eq.(\ref{eq:rdvz}) with respect to each of the twelve target stations. 

\begin{figure}[ht]
	\centering
     \includegraphics[width=.9\textwidth]{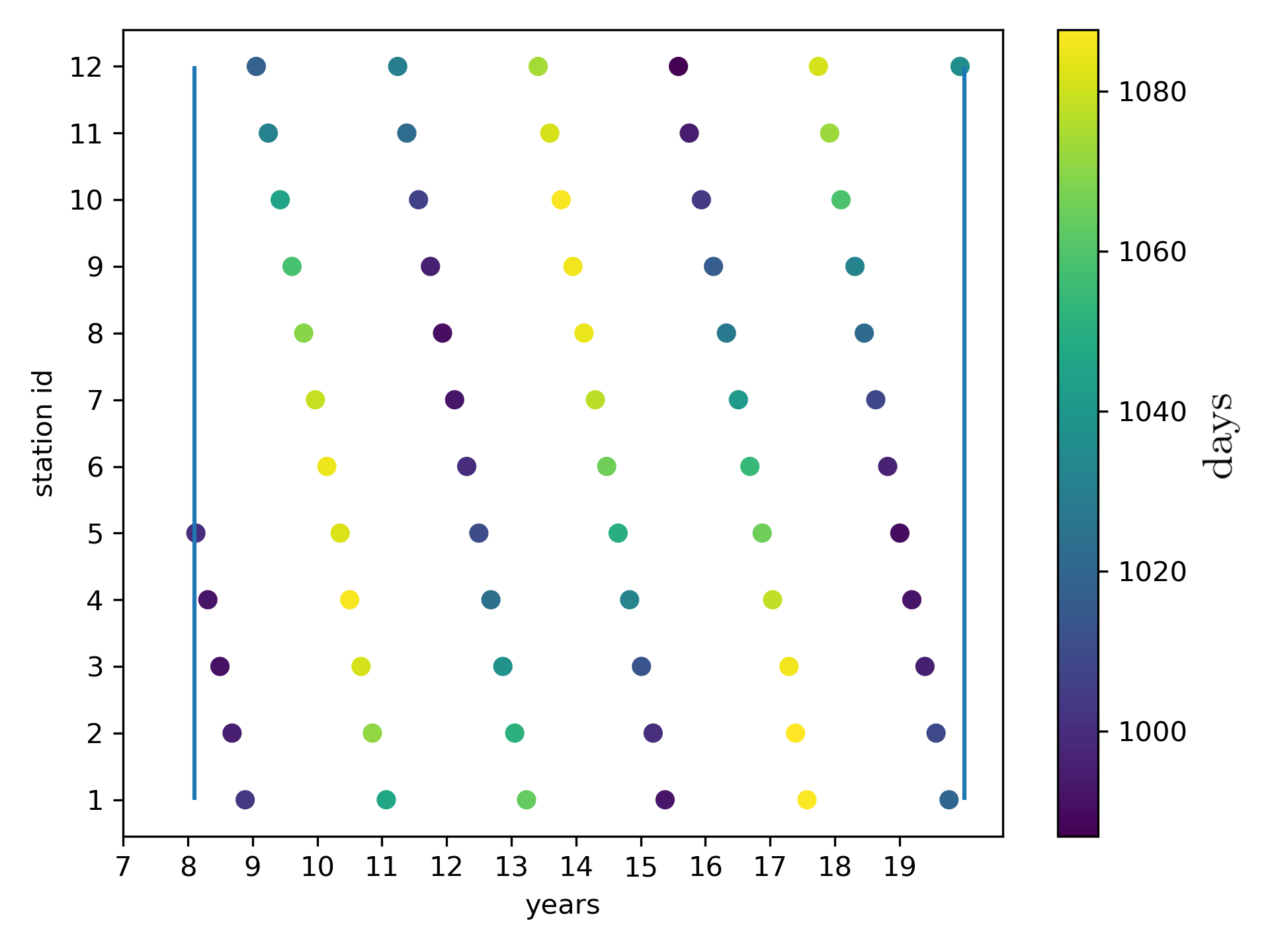}
    \caption{Each dot represents an optimally-phased transfer opportunity (marking its arrival time) for an example asteroid at $a=2.32$, $e=0.031$ and $i=1.14\degree$, earliest possible activation time $5.3$ years after mission start. The blue vertical lines mark the end of the mission and the earliest possible transfer opportunity within the mission time frame. The Edelbaum approximated transfer time for the particular asteroid shown is $T_{e,i} = 982.5$ days, while the actual minimum transfer time is $T_{min,i} = 990$ days and the highest is $T_{max,i} = 1088$ days. A brighter color indicates a qualitatively higher arrival mass at the corresponding station.}
    \label{fig:well_phased}     
\end{figure}


We will formally collect these transfers in a matrix $\mathbf M$, whose $i,j$ entry is a set $(t_k, m_k), k=1..K_{ij}$ containing all $K_{ij}$ epochs $t_k$ at which the mass $m_k$ from asteroid $i$ can be brought to the station $j$ using an optimally-phased transfer. To compute $\mathbf M$ efficiently, we make use of the following trick: In a first step, we find for asteroid $i$ a single optimally phased transfer to station $j=1$. This is done by restricting its arrival time window $[\underline t_f, \overline t_f]$ to the interval $[t_i^{act}, t_i^{act} + \delta T_{e,i}]$, using the corrected Edelbaum approximation $T_{e,i}$ presented in Section~\ref{subsec:edelbaum_learn} and some slack $\delta > 1$. We denote the time of this transfer by $T_i$. After determining $T_i$ by solving this OCP instance, all other transfers can be found by solving similar OCPs, which can now be constructed by exploiting an apparent periodicity in the solutions space. This periodicity is visualized for an example asteroid in Figure~\ref{fig:well_phased}. While the periodic structure depends on the orbital elements of each asteroid and becomes less linear and predictable for higher inclinations or eccentricities, it nevertheless is strong enough to infer excellent initial guesses for the solvers to use.

The initial guesses are constructed by computing the synodic period $T_{syn,i}$ between the orbit of asteroid $i$ and the Dyson ring orbit. Given the actual transfer time $T_i$ as solution from the first OCP, all other solutions from asteroid $i$ to the first station $j = 1$ can be found by adding multiples of $T_{syn,i}$ and bracketing the arrival window $[\underline t_f, \overline t_f]$ to narrow time intervals around these points:
\[
[t_i^{act} + T_i + k \cdot T_{syn,i} - \varepsilon, t_i^{act} + T_i + k \cdot T_{syn,i} + \varepsilon]
\]
for all $k$ that allow us to remain within the mission time. To find the transfers from $i$ to all other stations, we followed a similar procedure, by adding an additional offset of $j \cdot \frac{T_{syn,i}}{12}$ for every station $j = 2, \ldots, 12$ to the bracketing of the arrival window. 



\subsection{Ring Building Schedule}
\label{subsec:ring_buidling_schedule}

Following the preceding generation of $\mathbf M$, it remains to select for each asteroid, which of the many possible transfer opportunities it ultimately should take. This problem is closely related to the classical combinatorial assignment problem~\cite{burkard2012assignment}, where asteroids can be thought of as agents and stations as tasks. However, additional complexities arise in our case: the largest constraint is that only a single station can receive asteroids at a time. Formally, we define a set of receiving time windows $\mathcal{W} = \lbrace (W_\mathrm{1,begin}, W_\mathrm{1,end}), \ldots, (W_\mathrm{12,begin}, W_\mathrm{12,end}) \rbrace$ for each of the twelve stations, where for one station $j$ the earliest and the latest epoch at which it accepts incoming asteroids is described by $W_{j,\mathrm{begin}}$ and $W_{j,\mathrm{end}}$ correspondingly. Furthermore, we require that there is no overlap between any pair of time windows and that a gap of at least 90 days is left between temporally consecutive windows, as demanded by the problem formulation of the challenge.

Given that the $\Delta V_k^{total}$ and $a_D$ component of $J$ (overall objective function, see Eq.(\ref{eq:objfun})) are fixed at this point during the pipeline, the task of the ring building schedule remains to solve the assignment problem and determine a valid set of time windows $\mathcal{W}$ such that the last important component $M_{min}$ is maximized. For this purpose, we model the scheduling task as a bi-level optimization problem: The outer level problem encodes the intervals of a candidate window allocation $\mathcal{W}$ while the inner level solves the corresponding assignment problem using $\mathbf M$ constrained by $\mathcal{W}$.

For the outer level, we deploy a simple differential evolution solver to find optimal $\mathcal{W}$, guided by $M_{min}$ (the solution of the inner optimization problem) as an objective. For the inner level problem we deploy a modified Hungarian algorithm to find a solution to the specific instances of the constrained assignment problems. In the following, we describe the necessary steps of the latter procedure in detail.

\subsection{Inner Level Problem: Path-Based Refinement}
\label{subsec:path_based_refinement}
Given the matrix $\mathbf M$ and a time window allocation $\mathcal{W}$, we create a filtered matrix $\mathbf M'$ by only allowing transfers from $\mathbf M$ that arrive at their destination within the time windows defined by $\mathcal{W}$. Thus, an entry in $\mathbf{M'}_{ij}$ might be empty if for asteroid $i$ no arrival to station $j$ within $(W_{j,\mathrm{begin}}, W_{j,\mathrm{end}})$ is available. Should there be multiple arrivals available, we select the transfer that allows for a maximum mass to be delivered. More formally, $\mathbf{M'}_{ij} = \mathbf{M}_{ijk}$, with 

\begin{equation}
k = \argmax_{m_k}  ((t_k, m_k) \in \mathbf{M}_{ij} \mid W_{j,\mathrm{begin}} \leq t_k \leq W_{j,\mathrm{end}})
\end{equation}

Consequently, for each asteroid-station pair, at most one transfer opportunity is left in $\mathbf{M'}_{ij}$. The problem remains to decide however, which asteroid will go to which station, i.e. which of the asteroid-station pairs to select without violating the constraint that one asteroid may be only assigned to one station.

We model this problem as a weighted bipartite graph, with all asteroids on one side, all stations on the other side and an edge between an asteroid $i$ and a station $j$ if $i$ can reach $j$ with non-zero mass. The weight of the edge $ij$ is the mass that asteroid $i$ can deliver to $j$, as given by $\mathbf{M'}_{ij}$.

\begin{figure}
	\centering
    \subcaptionbox{\label{fig:hungarian-sketch-initial}}{\includegraphics[width=0.2\textwidth]{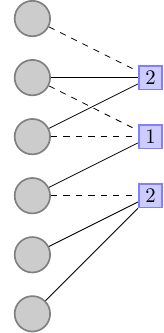}}
    \subcaptionbox{\label{fig:hungarian-sketch-path}}{\includegraphics[width=0.2\textwidth]{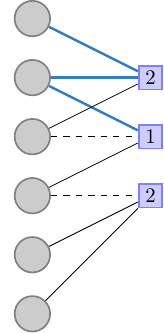}}
    \subcaptionbox{\label{fig:hungarian-sketch-optimized}}{\includegraphics[width=0.2\textwidth]{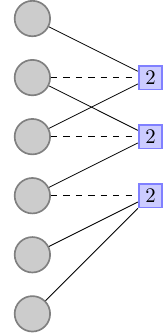}}
    \caption{Bipartite graph of a simple asteroid to station assignment problem. Six asteroids are assigned to three stations, each edge corresponds to a possible transfer. Solid lines signify a matched edge (i.e., a selected transfer), dashed lines an unmatched edge. Figure~\ref{fig:hungarian-sketch-initial} shows an unbalanced assignment: Station 2 has only one asteroid, and asteroid 1 is free but cannot be assigned to it, as it has no possible transfer. Figure~\ref{fig:hungarian-sketch-path} shows an augmenting path, drawn in thick blue lines, alternating matched and unmatched edges. Figure~\ref{fig:hungarian-sketch-optimized} shows the improved solution: Switching matched and unmatched edges on the augmenting path.}
    \label{fig:hungarian-sketch}
\end{figure}

Finding an optimal asteroid-to-station assignment is then equivalent to finding a matching in this graph which maximizes the mass arriving at the minimum mass station. Figure~\ref{fig:hungarian-sketch-initial} illustrates such an assignment graph for a simple example of six asteroids and three stations, assuming equal arrival masses for each transfer.

While for a classical assignment problem the sum of the weights of the matching is to be maximized, our objective is to maximize the minimum total weight over all stations. This difference makes the problem NP-hard, as the multiway number partitioning problem~\cite{doi:10.1137/0117039}, specifically maximizing the smallest sum, can be reduced in polynomial time to it.\footnote{Sketch of proof: Given a $k$-way number partitioning problem with a multiset $S$ of numbers, create $k$ building stations and one asteroid for each number $s_i \in S$, with a mass equal to $s_i$. For each value $m \in \mathbb{R}$, there is a partition of $S$ into $k$ subsets with sum at least $m$ exactly if there is an allocation of asteroids with the minimum-mass station having a mass of at least $m$.}

Consequently, finding an exact solution for large instances of the asteroid assignment problem is prohibitive. Thus, we deploy a heuristic inspired by the Hungarian algorithm~\cite{kuhn1955hungarian} for the assignment problem.

Given an assignment graph $G$, we define an alternating path in $G$ to be a connected sequence of edges, alternating between matched and unmatched edges. Figure~\ref{fig:hungarian-sketch-path} shows an alternating path of length $k=4$: station 2 $\leftrightarrow$ asteroid 2, $\leftrightarrow$ station 1 $\leftrightarrow$ asteroid 1. Switching all edges from assigned to unassigned and vice versa yields another valid assignment graph, shown in Figure~\ref{fig:hungarian-sketch-optimized}. We define the gain of a path with respect to the cost as the difference in value caused by applying it. Our approach for improving our assignment consists of repeatedly finding and applying augmenting paths with positive gain, as described in Algorithm~\ref{alg:hungarian}. 

\begin{algorithm}[ht!]
\caption{Path-based Refinement of asteroid-station allocation}
\label{alg:hungarian}
\begin{algorithmic}
\Require{transfer matrix  $\mathbf M$, time windows $\mathcal{W}$, maximum path length $k$, [initial assignment]}
\State $\mathbf{M'} \leftarrow$ filter transfers in $\mathbf M$ by windows $\mathcal{W}$
\State $G \leftarrow$ create assignment graph from $\mathbf{M'}$, setting all edges to unmatched
    \For{asteroid-station pair $(i, j)$ in initial assignment}
	\State mark edge in $G$ as matched
	\EndFor
\While{improvement found} \Comment{Main Loop}
	\For{station $s$} \Comment{Sorted by increasing mass}
		\State paths $\leftarrow$ generate alternating paths of length $k$, starting from $s$
		\State gains $\leftarrow$ calculate gains of all paths
	\EndFor
	\If{maximum gain $>$ 0}
	\State apply path of maximum gain to $G$
	\EndIf
\EndWhile
\State \Return assignment from $G$
\end{algorithmic}
\end{algorithm}

Regarding this algorithm, there a few things to note: First, since the number of paths of length $k$ grows exponentially, the running time is exponential in $k$ as well. In practice, we observed that the increase in running time for any $k>4$ does not justify the improvements in our objective function. We thus limit the path length $k$ for this step of our pipeline to $k=2$ for use in the differential evolution, and $k=4$ for post-processing refinements.
Secondly, the algorithm benefits from being bootstrapped by an initial assignment. We construct this initial assignment by deploying a greedy scheduling algorithm, described in Algorithm~\ref{alg:greedy_allocation}. Given an initial target mass $m$, the greedy scheduling incrementally increases the mass of all yet unconstructed stations by always selecting the earliest possible arrival opportunity. Once a station $j$ accumulates a mass larger than $m$, it is considered constructed and we can infer a construction time window $(W_{j,\mathrm{begin}}, W_{j,\mathrm{end}})$ by its earliest and latest arriving asteroid. Once a station $j$ is constructed, all asteroids arriving in its time window are fixed to the station and the accumulated masses in the remaining stations are set again to zero as only one station can be build at a time. After adding the gap of $90$ days to $W_{j,\mathrm{end}}$, we proceed to fill up the remaining stations using the earliest arrival of the yet unassigned asteroids. The algorithm terminates once all twelve stations are successfully constructed (in which case the minimum mass station has a mass of at least $m$) or if no more assignments can be added, potentially leaving stations unconstructed or with low accumulated mass. To avoid the latter outcome, the target mass $m$ needs to be set conservatively, i.e. by increasing $m$ in small amounts and restarting the greedy scheduler until the best greedy assignment is found.

\begin{algorithm}
\caption{Greedy scheduling}
\label{alg:greedy_allocation}
\begin{algorithmic}
\Require{transfer matrix $\mathbf M$, target average mass $m$}
\For{$(t_k, m_k) = \mathbf{M}_{ijk}$ chronologically}
	\State add $m_k$ to $mass(j)$
	\State add asteroid $i$ to $A(j)$
	\State advance time to $t_k$
	\If{unconstructed station $j$ with $mass(j) > m$ exists}
		\State fix asteroids $A(j)$ and construct station $j$
		\State determine $(W_{j,\mathrm{begin}}, W_{j,\mathrm{end}})$ by earliest and latest arriving asteroid
		\State reset $A(l) = \emptyset$ for all stations $l$ that still need construction
		\State reset $mass(l) = 0$ for all stations $l$ that still need construction
		\State exclude station $j$ and all asteroids $i \in A(j)$ from $\mathbf M$
		\State advance time to $W_{j,\mathrm{end}} + 90$d
	\EndIf
\EndFor
\State \Return assigned asteroids $A(j), j = 1, \ldots, 12$ and time windows $\mathcal{W}$
\end{algorithmic}
\end{algorithm}

\section{Results}

\subsection{Mother Ship Tree Searches}

Due to the time limitations of GTOC~11 we manually balanced the various hyper-parameters of our LRTS during experiments in order to quickly generate a large collection of feasible single mother ship trajectories of high $J'$. The most common choices for beam-width were $b \in \{10,20,30\}$, for fanout $g = \{5, 10, 20\}$ and $a_{D} \in \{1.2, 1.25, 1.3, 1.35, 1.40\}$ AU. The time of flight for each A2A-leg was bounded by $380$d and for the initial E2A-leg by $480$d. Out of all available asteroids, we computed the arrival mass $m_{arr}(a_D)$ for a fixed $a_{D}$ using the improved Edelbaum approximation and considered only asteroids above the $0.9$-quantile as valid targets. The length of a time-slice was fixed to $91.3125$d, which amounts to a total of $80$ time-slices for the whole mission duration. Executing the tree search in parallel on our servers and a high performance computation cluster allowed us to generate $145\,000$ raw single mother ship solutions, of which we selected the $7\,000$ best (according to $J'$) for further optimization.

\begin{figure}[h!]
	\centering
     \includegraphics[width=\textwidth]{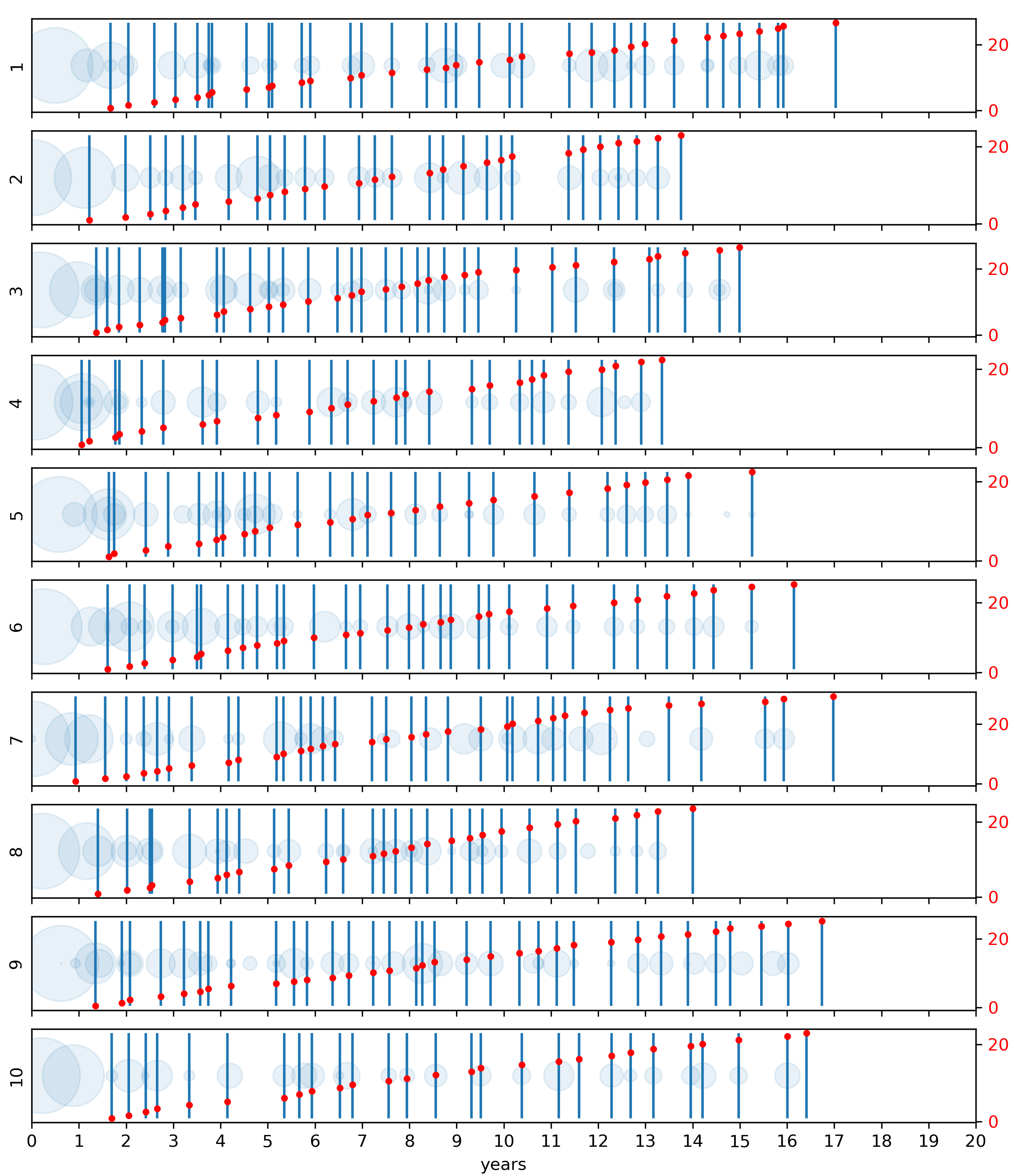}
    \caption{Ensemble of ten mother ship trajectories. The vertical lines correspond to asteroid activation epochs. The blue circles are plotted in correspondence to the $\Delta V$ manoeuvres epochs and have size proportional to the magnitude. The red dots mark the cumulative asteroid mass accumulated in units of the maximal mass $m_{max}$ of the dataset.} \label{fig:ensemble}     
\end{figure}

Given our approach of constructing the mother ship trajectories leg by leg using the simplified three-impulse model (see Section~\ref{subsec:a2a}), the $\Delta V$ requirements still had potential for improvement by local optimization starting from the full asteroid sequence (without altering it). For that purpose, we deployed the sequential least squares programming (SLSQP) solver from Python's scipy library for 10 iterations on each trajectory. Obtaining lower $\Delta V$ resulted in a new ordering of the $7\,000$ trajectories according to $J'$, which constitutes the set $\mathcal{U}$ for the following mother ship ensemble selection. Due to its randomness, we executed Algorithm~\ref{alg:greedy} multiple times to generate promising mother ship ensembles. Figure~\ref{fig:ensemble} shows the final ensemble, which was part of our highest scoring submission. In total, this ensemble collects $301$ asteroids with a DV-factor $\sum\limits_{k=1}^{10}\left(1 + \Delta V_{k}^{Total} / 50 \right)^2 \approx 19.1269$. 

\subsection{Ring Building Pipeline}

Following the steps of the pipeline, we determined a station ring radius of $a_D = 1.3198$ AU, an inclination of $i_D = 1.4282\degree$ and a right ascension of the ascending node $\Omega_D = 104.4039\degree$ by solving the optimization problem outlined in Eq.(\ref{eq:ringparameters}) using Co-variance Matrix Evolutionary Strategy (CMA-ES). The phase $\phi_1$ of the first station was not optimized and set to $0\degree$, resulting in a general phase of $\phi_j = 30\degree \cdot (j - 1)$ for all stations $j = 1, \ldots, 12$. 

In a next step, we constructed for each asteroid of the mother ship ensemble a first transfer to the first station $j=1$ as described in Section~\ref{subsec:optimally_phased_traj}. The resulting OCP (compare Eq.(\ref{eq:rdvz})) is solved by randomly initializing the Position and velocity co-states in $[-1, 1]$ and deploying SNOPT, which resulted in $301$ initial transfer solutions. Out of those, we construct the complete matrix $\mathbf M$ of all possible optimally-phased transfer opportunities for each asteroid to each station. Following this procedure, additional 14\,955 OCPs were generated and solved in parallel on our server equipped with 2xAMD EPYC ROME 64-CORE 7702 3.35GHZ CPUs, which can solve this amount of OCPs in less than one hour. The resulting transfer matrix $\mathbf M$ contained $15\,256$ transfer opportunities, approximately $4$ for each asteroid-station pair (more or less depending on the activation epoch and orbit of the asteroid in question).
 
Following the construction of the matrix $\mathbf M$, the greedy algorithm described in Algorithm~\ref{alg:greedy_allocation} was used to bootstrap the ring schedule optimization. Given $m_{max} = 1.963 \cdot 10^{14}$ kg to be the highest mass among all asteroids of the problem description, we assigned a starting mass of $m = 9m_{max}$ to the greedy algorithm and increased $m$ by steps of $\Delta m = 0.05m_{max}$. The highest $m$ for which the greedy scheduler still found an equally distributed balance was $m = 9.5m_{max}$. Given the assignment from the greedy scheduler and its time window allocation $\mathcal{W}$, the bi-level optimization problem as described in Section~\ref{subsec:ring_buidling_schedule} was constructed. The outer optimization problem was solved deploying an asychronous island model with three islands, each with a population of size $10$ initialized by $\mathcal{W}$. The modified Hungarian algorithm (Algorithm~\ref{alg:hungarian}) was restricted to paths of length $k=2$ for solving the inner optimization problems, returning $M_{min}$ as fitness. Thus, for the outer optimization problem, a differential evolution algorithm modified $\mathcal{W}$ by selecting the individuals of high $M_{min}$ in the population for recombination and propagation to the next evolutionary generation.

Since this fitness function is costly to evaluate (as it involves solving the asteroid assignment problem) we could run differential evolution for ten generations only, which nevertheless was sufficient to improve $M_{min}$. After this step, we applied some post-processing refinements to the best found solution of the assignment problem by running Algorithm~\ref{alg:hungarian} with path lengths of $k=3$ and $k=4$. Overall, the initial minimum station mass of $m = 9.5m_{max}$ constructed by the greedy scheduler could be improved to $m =  10.24m_{max}$ which is our final $M_{min}$. 


\begin{table}[h]
\centering
\begin{tabular}{|c|c|}
\hline
$a_D$ [AU]                  & $1.3198$                 \\ \hline
$i_D$ [$\deg$]              & $1.4282$                 \\ \hline
$\Omega_D$ [$\deg$]         & $104.4039$               \\ \hline
$M_{min}$ [kg]              & $2.0125 \cdot 10^{15}$   \\ \hline
$N_{tot}$                   & $301$                    \\ \hline
$J$ [-]                     & $6359.7249$              \\ \hline
\end{tabular}
\caption{Characteristics of the ACT\&Friends' final submitted solution to the GTOC~11 challenge, ranked second place in GTOC~11.}
\label{tab:finalsolution}
\end{table}

Everything taken together, the parameters of the solution that our pipeline produced are summarized in Table~\ref{tab:finalsolution}. Our server finished this computation in about 2-3 hours with the majority of time spent on OCP solving and the optimization loops for the scheduling problem. The computational cost depends largely on the parallelization capabilities of the hardware and is thus not easily quantifiable. Our overall score of $J = 6359.7249$ (not accounting for the bonus factor $B$) allowed us to place second in GTOC~11. A short animation showcasing the trajectories for the construction of our Dyson ring can be found online~\cite{ourvideo} alongside the winning solution of Tsinghua university.

\section{Conclusions}

The presented work of this report was developed with the goal in mind to obtain the highest possible rank during GTOC~11. Given the fiercely competitive nature of recent GTOCs, our team was forced to innovate on our techniques, which would not have been sufficient by themselves. One innovation was the addition of machine learning to improve accuracy and provide better informed decisions.

As numerous OCP problems are typically encountered during GTOCs but also for complex trajectory optimization tasks in general, it is of importance to utilize optimal solutions not only directly (i.e. to advance a combinatorial search) but also indirectly, by building databases for machine learning. Distilling a model to correct the error of the Edelbaum approximation for example enabled mayor improvements for our tree search algorithms, but was also essential for multiple independent steps of our ring building pipeline, due to the strong relation between $T_{e}$ and $M_{min}$.
Considering the difficult fitness landscapes of the OCP problems involved, rapid deployment of fast evolutionary optimization techniques are still essential for success. While linking small building blocks of evolutionary optimization is standard practice to solve combinatorial problems via tree searches or ant colony optimization, an additional innovation described in this report was the application of the modified Hungarian algorithm as part of an evolutionary outer loop. This provides a fusion of traditional assignment/scheduling algorithms with techniques from evolutionary optimization which turned out to be highly effective. 

Although we might still need a few more centuries to see something like the Dyson ring become a reality in our solar system, the current trends regarding infrastructure in space are already visible. Consequently, one may safely assume that scheduling and planning problems similar to the GTOC~11 challenge will only increase in relevance and thus provide a new and worthwhile opportunity to study trajectory optimization at larger scales.

\bibliography{references}

\end{document}